\documentclass[11pt,a4paper]{article}
\usepackage[hyperref]{emnlp2018}
\usepackage{times}
\usepackage{amsmath,amsthm,amssymb}
\usepackage[vlined]{algorithm2e}
\usepackage{latexsym}
\usepackage{hhline}
\usepackage{graphicx}
\usepackage{multirow}
\usepackage{url}
\usepackage{booktabs}
\usepackage{color,soul}
\usepackage{linguex}

\usepackage{enumitem}
\setitemize{noitemsep,topsep=0pt,parsep=0pt,partopsep=0pt}

\aclfinalcopy 
\def\aclpaperid{1474} 

\newcommand{\ignore}[1]{}

\definecolor{paleblue}{rgb}{0.69, 0.93, 0.93}
\newcommand{\hlc}[2][yellow]{{\sethlcolor{#1}\hl{#2}}}

\DeclareMathOperator*{\argmax}{arg\,max}
\newcommand*\samethanks[1][\value{footnote}]{\footnotemark[#1]}

\title{Learning To Split and Rephrase From Wikipedia Edit History}

\author{%
    Jan A. Botha\Thanks{ Both authors contributed equally.} \and
    Manaal Faruqui\samethanks \and
    John Alex \and
    Jason Baldridge \and
    Dipanjan Das \\[1ex]
    {\small \tt \{jabot,mfaruqui,jpalex,jasonbaldridge,dipanjand\}@google.com} \\[1ex]
    Google AI Language\\}
    
\date{}

\begin{document}
\maketitle
\begin{abstract}
  Split and rephrase is the task of breaking down a sentence into shorter ones that together convey the same meaning. We extract a rich new dataset for this task by mining Wikipedia's edit history: WikiSplit contains one million naturally occurring sentence rewrites, providing sixty times more distinct split examples and a ninety times larger vocabulary than the WebSplit corpus introduced by \newcite{narayan:2017} as a benchmark for this task. Incorporating \mbox{WikiSplit} as training data produces a model with qualitatively better predictions that score 32 BLEU points above the prior best result on the WebSplit benchmark.
\end{abstract}

\section{Introduction}
A complex sentence can typically be rewritten into multiple simpler ones that together retain the same meaning. Performing this \textit{split-and-rephrase} task is one of the main operations in text simplification, alongside paraphrasing and dropping less salient content
{\cite[i.a.]{siddharthan2006syntactic,zhu2010,woodsend-lapata2011}}.
The area of automatic text simplification has received a lot of attention \cite{siddharthan2014survey,shardlow2014survey}, yet still holds many open challenges \cite{Xu2015tacl-simplification}.
Splitting sentences in this way could also benefit systems where predictive quality degrades with sentence length, as observed in, e.g., relation extraction \cite{Zhang:position-aware-re:2017} and translation \cite{KoehnKnowles:challenges2017}.
And the schema-free nature of the task may allow for future supervision in the form of 
crowd-sourced rather than expensive expert annotation \cite{He:qasrl2015}.

\newcite{narayan:2017} introduce the WebSplit corpus for the split-and-rephrase task and report results for several models on it. \newcite{aharoni:2018} improve WebSplit by reducing overlap in the data splits, and demonstrate that neural encoder-decoder models \cite{bahdanau2014neural} perform poorly, even when enhanced with a copy mechanism \cite{Gu2016-copynet,See2017-ptr-generator}.

\begin{figure}[!tb]
    \centering
    \footnotesize{
    {\fontfamily{cmss}\selectfont 
        \begin{tabular}{p{0.9\linewidth}}
            A classic leaf symptom is water-soaked lesions between the veins \hlc{which appear as} angular \hlc{leaf-spots} where the lesion edge and vein meet.
            \\ 
            
            \noindent\hfil\rule[0.5ex]{0.7\linewidth}{0.5pt} \hfil
            
            A classic leaf symptom is \hlc[paleblue]{the appearance of angular,} water-soaked lesions between the veins\hlc[paleblue]{. The} angular \hlc[paleblue]{appearance results} where the lesion edge and vein meet.
            \\ 
        \end{tabular}  
    }
    }
    \caption{A split-and-rephrase example extracted from a Wikipedia edit, where the top sentence had been edited into two new sentences by removing some words (yellow) and adding others (blue).
    }
    \label{fig:wiki-split-extraction-example}
\end{figure}

One limitation of the WebSplit examples themselves is that they contain fairly unnatural linguistic expression using a small vocabulary. We introduce new training data mined from Wikipedia edit histories that have some noise, but which have a rich and varied vocabulary over naturally expressed sentences and their extracted splits. Figure \ref{fig:wiki-split-extraction-example} gives an example of how a Wikipedia editor rewrote a single sentence into two simpler ones. We create WikiSplit, a set of one million such examples mined from English Wikipedia, and show that models trained with this resource produce dramatically better output for split and rephrase.

Our primary contributions are:
\begin{itemize}
\item A scalable, language agnostic method for extracting split-and-rephrase rewrites from Wikipedia edits.
\item Public release of the English WikiSplit dataset, containing one million rewrites: {\small \url{http://goo.gl/language/wiki-split}}
\item By incorporating WikiSplit into training, we more than double (30.5 to 62.4) the BLEU score obtained on WebSplit by \newcite{aharoni:2018}.
\end{itemize}

\begin{table*}[!t]
\small
  \centering
  \begin{tabular}{p{0.97\linewidth}}
  \multicolumn{1}{l}{\emph{Correct}} \\ \midrule 
  \emph{Street Rod is the first in a series of two games released for the PC and Commodore 64 in 1989.}\\
   Street Rod is the first in a series of two games. 
   It was released for the PC and Commodore 64 in 1989.\\
   \cmidrule(lr){1-1}
  \emph{He played all 60 minutes in the game and rushed for 114 yards, more yardage than all the Four Horsemen combined.}\\
   He played all 60 minutes in the game.
   He rushed for 114 yards, more yardage than all the Four Horsemen combined.\\
  \cmidrule{1-1}
  \multicolumn{1}{l}{\emph{Unsupported}} \\ \midrule 
  \emph{When the police see Torco's injuries, they send Ace to a clinic to be euthanized, but he escapes and the clinic worker covers up his incompetence.}\\
   When the police see Torco's injuries \hlc[paleblue]{to his neck, they believe it is a result of Ace biting him}.
   They send Ace to a clinic to be euthanized, but he escapes and the clinic worker covers up his incompetence.\\
  \cmidrule{1-1}
  \multicolumn{1}{l}{\emph{Missing}} \\ \midrule 
  \emph{The avenue was extended to Gyldenl{\o}vesgade by Copenhagen Municipality in 1927-28 and \hlc{its name was changed to Rosen{\o}rns All{\'e}} after Ernst Emil Rosen{\o}rn (1810-1894)} .\\
   The avenue was extended to Gyldenl{\o}vesgade by Copenhagen Municipality in 1927-28.
   The street was named after Ernst Emil Rosen{\o}rn (1810-1894) .\\
  \midrule
  \end{tabular}
  \caption{Examples of correct and noisy sentence splits extracted from Wikipedia edits. Noise from unsupported or missing statements is visualized with the same coloring as in Figure~\ref{fig:wiki-split-extraction-example}.
  }
  \label{tab:wiki-split-examples}
\end{table*}

\section{The WikiSplit Corpus}

WebSplit provides a basis for measuring progress on splitting and rephrasing sentences.
However, its small size, inherent repetitiveness, and synthetic nature limit its broader applicability. In particular, we see it as a viable benchmark for evaluating models, but not for training them. To that end, we introduce the WikiSplit corpus and detail its construction next.

\subsection{Mining Wikipedia Edits}

Wikipedia maintains snapshots of entire documents at different timestamps, which makes it possible to reconstruct edit histories for documents. This has been exploited for many NLP tasks, including sentence compression \cite{Yamangil-2008}, text simplification \cite{Yatskar-2010,woodsend-lapata2011,tonelli2016simpitiki} and modeling semantic edit intentions \cite{yang-EtAl:2017:EMNLP20174}.

To construct the WikiSplit corpus, we identify edits that involve sentences being split. A list of sentences for each snapshot is obtained by stripping HTML tags and Wikipedia markup and running a sentence break detector \cite{gillick:2009}. Temporally adjacent snapshots of a Wikipedia page are then compared to check for sentences that have undergone a split like that shown in Figure~\ref{fig:wiki-split-extraction-example}. We search for splits in both temporal directions.

Given all candidate examples extracted this way, we use a high-precision heuristic to retain only high quality splits. To extract a full sentence $C$ and its candidate split into $S=(S_1,S_2)$, we require that $C$ and $S_1$ have the same trigram prefix, $C$ and $S_2$ have the same trigram suffix, and $S_1$ and $S_2$ have different trigram suffixes.
To filter out misaligned pairs, we use BLEU scores \cite{papineni:2002} to ensure similarity between the original and the split versions.

Specifically, we discard pairs where BLEU($C$, $S_1$) or BLEU($C$, $S_2$) is less than $\delta$ (an empirically chosen threshold).
If multiple candidates remain for a given sentence $C$, we retain $\argmax_S\left(\textrm{BLEU}(C,S_1)+\textrm{BLEU}(C,S_2)\right)$.%
\footnote{We attempted to mitigate other noise inherent in Wikipedia by removing items that 1) repeated a token more than three times in a row; 2) contained a token longer than 25 characters; 3) were suggestive of profane vandalism.}


\begin{table}[!tb]
    \centering
    \begin{tabular}{c|rrr|r}
    \hline
    \emph{Thresh.} $\delta$ & Correct & Unsupp. & Miss. & Size \\ \hline
        0.1     & 161  & 35  & 6 & 1.4m\\
        0.2     & 168  & 35  & 4 & 1.0m\\
        0.3     & 169  & 31  & 4 & 0.5m\\\hline
    \end{tabular}
    \caption{Quality vs corpus size trade-off when setting the similarity threshold. The counts are for a random sample of 100 split-and-rephrase examples extracted using our method (i.e., 200 simple sentences). Keys: \textbf{Unsupp}orted; \textbf{Miss}ing}
    \label{tab:corpus-sample-quality}
\end{table}

\subsection{Corpus Statistics and Quality}

Our extraction heuristic is imperfect, so we manually assess corpus quality using the same categorization schema proposed by \newcite{aharoni:2018}; see Table~\ref{tab:wiki-split-examples} for examples of \textit{correct}, \textit{unsupported} and \textit{missing} sentences in splits extracted from Wikipedia. We do this for 100 randomly selected examples using three different thresholds of $\delta$. As shown in Table~\ref{tab:corpus-sample-quality}, $\delta$=0.2 provides the best trade-off between quality and size.

Out of the 100 complex sentences in the sample, only 4 contained information that was not completely covered by the simple sentences.
In our corpus, every complex sentence is split into two simpler sentences, so the sample contains 200 simple sentences.
Out of these we found 168 (84\%) to be correct, while 35 (18\%) contained unsupported facts.
Thus, for the overall sample of 100 split-and-rephrase examples, 68\% are perfect while 32\% contain some noise (either unsupported facts or missing information). 
We stress that our main goal is to use data extracted this way as training data and accept that its use for evaluation is an imperfect signal with some inherent noise and bias (by construction).

\begin{table}[tb] 
    \centering
    \begin{tabular}{lrrrr}
         & \multicolumn{2}{c}{WebSplit} 
         & \multicolumn{2}{c}{WikiSplit}  \\
                 \cmidrule(lr){2-3} \cmidrule(lr){4-5} 
         & \multicolumn{1}{c}{Count} & \multicolumn{1}{c}{Unique} & \multicolumn{1}{c}{Count} & \multicolumn{1}{c}{Unique} \\
         \cmidrule(lr){2-2} \cmidrule(lr){3-3} \cmidrule(lr){4-4}
         \cmidrule(lr){5-5} 
    $C$  & 1.3m   & 17k & 1.0m    & 1.0m \\
    $S'$ & 6.1m   & 28k & 2.0m    & 1.9m\\
    $t$  & 344k   & 7k  & 33.1m	  & 633k \\
    \midrule
    \end{tabular}
    \caption{Training corpus statistics in terms of complex sentences ($C$), simple sentences ($S'$=$\cup_iS_i$) and tokens~($t$, appearing across unique complex sentences). WikiSplit provides much greater diversity and scale.
    }
    \label{tab:corpus-stats}
\end{table}

After extraction and filtering, we obtain over one million examples of sentence splits from around 18 million English documents.
We randomly reserved 5000 examples each for tuning, validation and testing, 
producing 989,944 unique complex training sentences,
compared to the 16,938 of WebSplit (cf. Table~\ref{tab:corpus-stats}).


\subsection{Comparison to WebSplit} 

\newcite{narayan:2017} derived the WebSplit corpus by matching up sentences in the WebNLG corpus \cite{gardent:2017} 
according to partitions of their underlying meaning representations (RDF triples). The WebNLG corpus itself was created by having crowd workers write sentential realizations of one or more RDF triples. The resulting language is often unnatural, 
for example,
\emph{``Akeem Dent once played for the Houston Texans team which is based in Houston in Texas.''}%
\footnote{Given RDF triple: \{(H\_Txns, city, Texas),
    (Akeem\_Dent, \mbox{formerTeam}, H\_Txns),
    (H\_Txns, city, Houston)\}.}

%

Repetition arises because the same sentence fragment may appear in many different examples. This is to be expected given that WebSplit's small vocabulary of 7k words must account for the 344k tokens that make up the distinct complex sentences themselves.
\footnote{We use {\bf WebSplit~v1.0} throughout, which is the scaled-up re-release by \newcite{narayan:2017} at
\url{http://github.com/shashiongithub/Split-and-Rephrase}, commit
\texttt{a9a288c}.
Preliminary experiments showed the same trends on the smaller v0.1 corpus, as resplit by \newcite{aharoni:2018}.}

This is compounded in that each sentence contains a named entity by construction.
In contrast, our large new \mbox{WikiSplit} dataset offers more natural and diverse text (see examples in Table~\ref{tab:wiki-split-examples}),
having a vocabulary of 633k items covering the 33m tokens in its distinct complex sentences. 

The task represented by our WikiSplit dataset is \emph{a priori} both harder and easier than that of the WebSplit dataset --
harder because of the greater diversity and sparsity, but potentially easier due to the uniform use of a single split.

Of the two datasets, WebSplit is better suited for evaluation: its construction method guarantees cleaner data than is achieved by our extraction heuristic, and it provides
multiple reference decompositions for each complex sentence, which tends to improve the correlation of automatic metrics with human judgment in related text generation tasks \cite{ToutanovaEtAl2016-compression}.

\section{Experiments}
In order to understand how WikiSplit can inform the split-and-rephrase task, we vary the composition of the training set when training a fixed model architecture. We compare three training configurations: \textsc{WebSplit} only, \textsc{WikiSplit} only, and \textsc{Both}, which is simply their concatenation.

Text-to-text training instances are defined as all the unique pairs of $(C,S)$, where $C$ is a complex sentence and $S$ is its simplification into multiple simple sentences \cite{narayan:2017,aharoni:2018}.
For training, we delimit the simple sentences with a special symbol.
We depart from the prior work by only using a subset of the WebSplit training set: we take a fixed sub-sample such that each distinct $C$ is paired with a single $S$, randomly selected from the multiple possibilities in the dataset.
This scheme produced superior performance in preliminary experiments.

As a quality measure, we report multi-reference \emph{corpus-level} BLEU\footnote{Using NLTK v3.2.2, with case sensitive scoring.} \cite{papineni:2002}, but include sentence-level BLEU (sBLEU) for direct comparison to past work.%
\footnote{Past work on WebSplit \cite{narayan:2017,aharoni:2018} reported macro-averaged \emph{sentence-level} BLEU, calculated without smoothing precision values of zero. We found this ill-defined case occurred often for low-quality output.}
We also report length-based statistics to quantify splitting.

We use the same sequence-to-sequence architecture that produced the top result for \newcite{aharoni:2018}, ``\mbox{Copy512}'', which is a one-layer, bi-directional LSTM (cell size 512) with attention \cite{bahdanau2014neural} and a copying mechanism \cite{See2017-ptr-generator} that dynamically interpolates the standard word distribution with a distribution over the words in the input sentence.
Training details are as described in the Appendix of \newcite{aharoni:2018} using the OpenNMT-py framework \cite{opennmt:2017}.%
\footnote{\href{http://github.com/OpenNMT/OpenNMT-py}{\texttt{github.com/OpenNMT/OpenNMT-py}}, \texttt{0ecec8b}}

\subsection{Results}
We compare to the \textsc{Source} baseline, which is the previously reported method of taking the unmodified input sentence as prediction,
and we add \textsc{SplitHalf}, the natural baseline of deterministically splitting a complex sentence into two equal-length token sequences and appending a period to the first one.

\begin{table}[!tb] 
    \centering
    \begin{tabular}{l|rr}
    $\downarrow$train/eval$\rightarrow$ & WebSplit 1.0 & WikiSplit \\ \hline
    \textsc{Source}        & 58.0 & 73.4  \\
    \textsc{SplitHalf}     & 54.9 & 71.7  \\ \hline
    \textsc{WebSplit}      & 35.3 & 4.2   \\
    \textsc{WikiSplit}     & 59.4 & 76.0 \\
    \textsc{Both}          & \textbf{61.4} & \textbf{76.1} \\\hline
    \end{tabular}
    \caption{Corpus-level BLEU scores on the validation sets for the same model architecture trained on different data.}
    \label{tab:dev-results}
\end{table}

Table \ref{tab:dev-results} compares our three training configurations on the validation sets of both WebSplit and WikiSplit.
The \textsc{WebSplit} model scores 35.3 BLEU on the WebSplit validation set but fails to generalize beyond its narrow domain, as evidenced by reaching only 4.2 BLEU on the \mbox{WikiSplit} validation set.

The example predictions in Table~\ref{tab:output-examples} illustrate how this model tends to drop content (``Alfred Warden'', ``mouth'', ``Hamburg''), hallucinate common elements from its training set (``food'', ``ingredient'', ``publisher'') and generally fails to produce coherent sentences.

\begin{table}[!tb]
    \centering
    \begin{tabular}{l|rrrr}
      & BLEU & sBLEU & \#S/C & \#T/S   \\ \hline
    Reference                   &   & --   & 2.5 & 10.9 \\ \hline
    \textsc{Source}             & 58.7 & 56.1 & 1.0 & 20.5 \\ 
    \textsc{SplitHalf}          & 55.7 & 53.0 & 2.0 & 10.8 \\ \hline
    \textsc{AG18}               & 30.5 & 25.5 & 2.3 & 11.8 \\ \hline
    \textsc{WebSplit}           & 34.2 & 30.5 & 2.0 & 8.8  \\
    \textsc{WikiSplit}          & 60.4 & 58.0 & 2.0 & 11.2 \\
    \textsc{Both}               & \textbf{62.4} & \textbf{60.1} & 2.0 & 11.0 \\\hline
    
    \end{tabular}
    \caption{Results on the WebSplit v1.0 test set when varying the training data while holding model architecture fixed: corpus-level BLEU, sentence-level BLEU (to match past work), simple sentences per complex sentence, and tokens per simple sentence (micro-average).
    \textsc{AG18} is the previous best model by \newcite{aharoni:2018}, which used the full WebSplit training set, whereas we downsampled it.
    }
    \label{tab:model-comparison}
\end{table}

\begin{table*}[tb]
    \centering
    \begin{tabular}{l|cccc}
    \emph{Training}    & Unsupported & Missing & Repeated & Correct \\ \hline
    \textsc{AG18}      & 82      & 45    & 12   & 26/119 (22\%) \\ \hline
    \textsc{WebSplit}  & 58      & 47    & 13   & 32/100 (32\%) \\
    \textsc{WikiSplit} & 8       & 5     & 0    & 91/100 (91\%) \\
    \textsc{Both}      & \textbf{4}       & \textbf{4}     & \textbf{0}    & \textbf{95/100 (95\%)} \\ \hline
    \end{tabular}
    \caption{Manual evaluation results, as counts over the simple sentences predicted by each model for a random sample of 50 inputs from WebSplit 1.0 validation set.}
    \label{tab:manus-evaluus}
\end{table*}

\begin{table*}[!tb]
  \centering
  \begin{tabular}{lp{0.8\linewidth}}
 
    \multicolumn{2}{l}{\textbf{Example 1} from WebSplit validation set}  \\
    \midrule
    \emph{Input} &
        \emph{Alan Bean , born in Wheeler , Texas , was a crew member of Apollo 12 where Alfred Worden served as the backup pilot~.} 
        \\ 
    
    \textsc{AG18} &
        Alan Bean is a member of the crew of Apollo 12 which was operated by backup pilot . Alan Shepard was born in Texas~.
        \\ 
        
    \textsc{WebSplit} &
        Alan Bean served as a crew member of Apollo 12 . Alan Bean was born in Wheeler~.
        \\ 
     
    \textsc{WikiSplit} &
        Alan Bean , born in Wheeler , Texas . He was a crew member of Apollo 12 where Alfred Worden served as the backup pilot~.
        \\
    
    \textsc{Both} &
        Alan Bean was born in Wheeler , Texas~. Alan Bean was a crew member of Apollo~12 . Alfred Worden served as the backup pilot~.
        \\
    \midrule
    \multicolumn{2}{l}{\textbf{Example 2} from WikiSplit validation set}  \\ 
    \midrule
    \emph{Input} &
        \emph{He is a short , dark grayish - green monster with a beak - like mouth and a cap made out of the upper part of a creature 's skull .}
        \\ 
    
    \textsc{WebSplit} &
        He is a food found in grayish . grayish monster is an ingredient in grayish .
        \\ 
     
    \textsc{WikiSplit} &
        He is a short , dark grayish - green monster with a beak - like mouth . A cap made out of the upper part of a creature 's skull .
        \\ 
    
    \textsc{Both} &
        He is a short , dark grayish - green monster with a beak - like mouth . His cap is made out of the upper part of a creature 's skull .
        \\  
        
    \midrule
    \multicolumn{2}{l}{\textbf{Example 3} from WikiSplit validation set}  \\
    \midrule
    \emph{Input} &
        \emph{Der begl{\"u}ckte Florindo is an opera composed by Handel in Hamburg in 1708 as part of a double opera , the other part being Die verwandelte Daphne .}
        \\
    
    \textsc{WebSplit} &
        Handel opera is the publisher of the opera opera . Handel is the capital of 1708 .
        \\ 
     
    \textsc{WikiSplit} &
        Der begl{\"u}ckte Florindo is an opera composed by Handel in Hamburg in 1708 . It was part of a double opera , the other part being Die verwandelte Daphne .
        \\ 
    
    \textsc{Both} &
        Der begl{\"u}ckte Florindo is an opera composed by Handel in Hamburg in 1708 as part of a double opera . The other part being Die verwandelte Daphne .\\
        \midrule
  \end{tabular}
  \caption{Example model predictions for items from each validation set. \textsc{AG18} gives the output of the Copy512-model of \newcite{aharoni:2018}, while the other outputs are from our models trained on the corresponding data.}
  \label{tab:output-examples}
\end{table*}

In contrast, the \textsc{WikiSplit} model achieves 59.4 BLEU on the WebSplit validation set, \emph{without observing any in-domain data}. It also outperforms the two deterministic baselines on both validation sets by a non-trivial BLEU margin.
This indicates that the WikiSplit training data enable better generalization than when using WebSplit by itself.
Reintroducing the downsampled, in-domain training data (\textsc{Both}) further improves performance on the WebSplit evaluation.


These gains in BLEU from using WikiSplit carry over to the blind manual evaluation we performed on a random sample of model predictions on the WebSplit validation set.
As shown in Table~\ref{tab:manus-evaluus},
the \textsc{Both} model produced the most accurate output (95\% correct simple sentences), with the lowest incidence of missed or unsupported statements.
Our manual evaluation includes the corresponding outputs from \newcite{aharoni:2018} (\textsc{AG18}), which were 22\% accurate.

The examples in Table~\ref{tab:output-examples} demonstrate that the \textsc{WikiSplit} and \textsc{Both} models produce much more coherent output which faithfully rephrases the input. In Example~1, the combined model (\textsc{Both}) produces three fluent sentences, overcoming the strong bias toward two-sentence output inherent in the majority of its training examples.

We relate our approach to prior work on WebSplit~v1.0 by reporting scores on its test set in Table~\ref{tab:model-comparison}.
Our best performance in BLEU is again obtained by combining the proposed WikiSplit dataset with the downsampled WebSplit, yielding a 32~point improvement over the prior best result.

\section{Conclusion and Outlook}
Our results demonstrate a large, positive impact on the split-and-rephrase task when training on large, diverse data that contains some noise. This suggests that future improvements may come from finding other such sources of data as much as from modeling. The new WikiSplit dataset is intended as training data, but for further progress on the split-and-rephrase task, we ideally need evaluation data also derived from naturally occurring sentences, and an evaluation metric that is more sensitive to the particularities of the task.

\section{Acknowledgments}
 Thanks go to Kristina Toutanova and the anonymous reviewers for helpful feedback on an earlier draft, and to Roee Aharoni for supplying his system's outputs. 

\vfill

\bibliography{emnlp2018}
\bibliographystyle{acl_natbib_nourl}



\end{document}